\documentclass[twoside,12pt,english,nontitlepage,openany]{article}
\usepackage{babel}
\usepackage[utf8]{inputenc}
\usepackage{color}
\definecolor{marron}{RGB}{60,30,10}
\definecolor{darkblue}{RGB}{0,0,80}
\definecolor{lightblue}{RGB}{80,80,80}
\definecolor{darkgreen}{RGB}{0,80,0}
\definecolor{darkgray}{RGB}{0,80,0}
\definecolor{darkred}{RGB}{80,0,0}
\definecolor{shadecolor}{rgb}{0.97,0.97,0.97}
\usepackage{graphicx}
\usepackage{wallpaper}
\usepackage{wrapfig}
\usepackage{booktabs}
\usepackage{amsmath}

\usepackage{fancyhdr}
\usepackage{lettrine}
\input Acorn.fd
\newcommand*\initfamily{\usefont{U}{Acorn}{xl}{n}}

\usepackage{geometry}
\usepackage[
final,
stretch=10,
protrusion=true,
tracking=true,
spacing=on,
kerning=on,
expansion=true]{microtype}

\usepackage{fourier-orns}

\newcommand{\ornamento}{\vspace{2em}\noindent \textcolor{darkgray}{\hrulefill~ \raisebox{-2.5pt}[10pt][10pt]{\leafright \decofourleft \decothreeleft  \aldineright \decotwo \floweroneleft \decoone   \floweroneright \decotwo \aldineleft\decothreeright \decofourright \leafleft} ~  \hrulefill \\ \vspace{2em}}}
\newcommand{\ornpar}{\noindent \textcolor{darkgray}{ \raisebox{-1.9pt}[10pt][10pt]{\leafright} \hrulefill \raisebox{-1.9pt}[10pt][10pt]{\leafright \decofourleft \decothreeleft  \aldineright \decotwo \floweroneleft \decoone}}}
\newcommand{\ornimpar}{\textcolor{darkgray}{\raisebox{-1.9pt}[10pt][10pt]{\decoone \floweroneright \decotwo \aldineleft \decothreeright \decofourright \leafleft} \hrulefill \raisebox{-1.9pt}[10pt][10pt]{\leafleft}}}



\newcommand{\estcab}[1]{\itshape\textcolor{marron}{\nouppercase #1}}

\usepackage{lipsum}
\usepackage[hang,splitrule]{footmisc}
\usepackage{biblatex}
\usepackage{chngcntr}

\bibliography{sources}

\title{Giving Up Control\\ Neurons as Reinforcement Learning Agents}
\author{Jordan Ott}
\date{}

\begin{document}
\maketitle
\ornamento

\lettrine[lines=3]{\initfamily\textcolor{darkgreen}{A}}{rtificial Intelligence} has historically relied on planning, heuristics, and handcrafted approaches designed by \textit{experts}. All the while claiming to pursue the creation of \textit{Intelligence}. This approach fails to acknowledge that intelligence emerges from the dynamics within a complex system. Neurons in the brain are governed by local rules, where no single neuron, or group of neurons, coordinates or controls the others. This local structure gives rise to the appropriate dynamics in which intelligence can emerge. Populations of neurons must compete with their neighbors for resources, inhibition, and activity representation. At the same time, they must cooperate, so the population and organism can perform high-level functions. To this end, we introduce modeling neurons as reinforcement learning agents. Where each neuron may be viewed as an independent actor, trying to maximize its own self-interest. 
By framing learning in this way, we open the door to an entirely new approach to building intelligent systems.

\ornamento

\section*{Key Points}
\begin{enumerate}
    \item We frame individual neurons as reinforcement learning agents
    \item Each neuron attempts to maximize its own rewards
    \item Neurons compete with their neighbors while simultaneously cooperating
    \item Networks of independent agents solve high-level reinforcement learning tasks
    \item While maximizing their own self-interest neurons learn to cooperate for the betterment of the collective 
    \item Biologically motivated - local - reward schemes improve network dynamics
    \item Biological exactness regarding ions, channels, and proteins are not as important as the competition, rewards, and computation paradigms they induce 
\end{enumerate}
\newpage

\section{Introduction}
\lettrine[lines=3]{\initfamily\textcolor{darkgreen}{E}}{ver} since John McCarthy coined the term, Artificial Intelligence has followed a single path. In order to create intelligent machines, algorithms are built on human-designed features believed to be important by those who constructed them. 

We cannot define intelligence, much less quantify it. This knowledge problem \cite{ott2020hayek} leads researchers to discretize the attributes of human intelligence - speech, vision, audition, planning, search. For each behavior humans can do a corresponding model is proposed \cite{ott2019questions}. These \textit{solutions} come in the form of handcrafted templates \cite{papert1966summer}, heuristics \cite{evans1964heuristic}, knowledge bases \cite{lenat1995cyc}, and the latest addition cost functions \cite{krizhevsky2012imagenet, ott2018learning, ott2018deep}. All in the quest for human-level intelligence. 

Deep learning has been successful because of its ability to approximate arbitrary functions \cite{cybenko1989approximation}. This paradigm of approximation has been extended to vision, speech, audition, and a host of other domains. In these settings, inputs are paired with target labels and the network seeks to minimize a cost function by making its predictions closer to the targets. Vision, for example, has been approximated by leveraging large labeled datasets of images. Where the information about an object's size, location, and type is available as the target label. This training data is used to minimize the corresponding cost function.

The performance of these networks on tasks like ImageNet is convincing, as they have \textit{learned} to identify thousands of object categories with superhuman performance. However, many shortcomings are evident, from a lack of common sense \cite{marcus2018deep}, to adversarial attacks \cite{su2019one}, and a failure to transfer knowledge \cite{recht2019imagenet}. These ailments prove behavior-level approximation yields the appearance of intelligence without understanding.

Building deeper models with more massive datasets \cite{shoeybi2019megatron, raffel2019exploring} is likely not the way to human-level intelligence. Nor will it come from more well-designed cost functions, regularization techniques, or efficient optimizers. The historical path of Artificial Intelligence - including deep learning - stands diametrically opposed to the paradigm from biology. To make progress, we must take note of the computational principles leveraged by the mammalian cortex. 

Top-down approaches like approximating behaviors - minimizing cost functions - fall critically short of how behaviors arise out of intra-cortical dynamics. The problem with behavior-level approximation is that it completely abstracts the underlying dynamics. By abstracting the behavior with a cost function, critical information is likely to be lost. Intelligence does not arise from behavior-level cost functions, symbolic representations, or any other abstraction. Intelligence emerges, bottom-up, from interactions between neurons, where each neuron is governed by local rules and acts according to its interests. 

Many pitfalls of current AI methods come from attempting to design agents for specific tasks. To diverge from this path, we propose framing learning as a multi-player competitive and cooperative game. Populations of neurons must compete with their neighbors for resources, inhibition, and activity representation. At the same time, they must cooperate, so the population and organism can perform high-level functions. To this end, we introduce modeling neurons as reinforcement learning agents. Each neuron may is modeled as an independent actor, trying to maximize its own self-interest.

\section{Related Works}
\begin{figure}
    \centering
    \includegraphics[width=\linewidth]{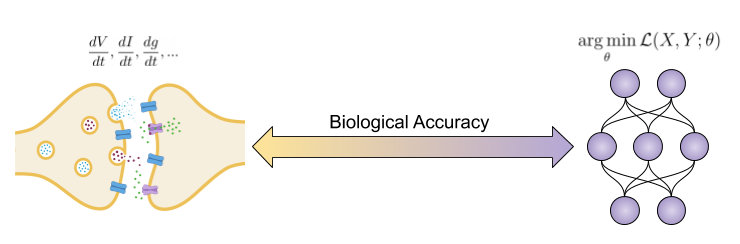}
    \caption{\footnotesize Neural models lie on a spectrum of anatomical accuracy. Left: At one end, there is strict adherence to biological details. Where models track changes in voltage, current, conductance, as well as a host of ion and receptor-specific details. Right: At the other end, there is less attention on individual neurons and systems are modeled abstractly by minimizing cost functions.}
    \label{fig:bio_spectrum}
\end{figure}

\lettrine[lines=3]{\initfamily\textcolor{darkgreen}{C}}{omputational models} lie on a spectrum of biological accuracy (Figure \ref{fig:bio_spectrum}). Trade offs in biological detail, efficiency, and practical applicability are made as one moves across this spectrum. 

On one end lies models that account for a variety of ions, molecules, and protein interactions within cells (Figure \ref{fig:bio_spectrum} left). Often these models are so complex that computational requirements restrict the effective size of the model. At this end, topics include modeling protein interactions within cells or ion flows through gated channels. These models provide us with near-exact dynamics of microscopic interactions. However, the sacrifice is made for computational speed and scalability. Large networks of these highly intricate neurons cannot run in real-time and are not suitable for real-world tasks - image recognition or autonomous vehicles. 

At the other end of the spectrum are models abstracting many, if not all, of the details regarding ions and neurotransmitters (Figure \ref{fig:bio_spectrum} right). These models sacrifice biological rigor for efficient solutions that scale well to large models. Examples at this end of the spectrum are current deep learning architectures. Models at this end rely on extreme abstractions of neurons to favor computational efficiency and increased performance on high level, behavioral tasks. 

As one moves from biological exactness to the more abstract, there is less emphasis on individual neurons. The focus becomes increasingly placed on behavior-level tasks where the network is evaluated on corresponding metrics \cite{ott2019questions}. In deep learning, this takes the form of minimizing cost functions. Reinforcement learning expresses this in reward prediction errors and temporal difference learning \cite{ljungberg1992responses, schultz1993responses, sutton1988learning}. Specifically, concerning deep learning, behaviors are approximated via direct cost functions. While this approach yields efficient solutions with adequate performance, it is likely to miss local interactions between neurons \cite{ott2019questions, ott2020hayek}. 

While some have described their work as ``reinforcement learning applied to single neurons" \cite{wang2015reinforcement} or ``neuron as an agent" \cite{ohsawa2018neuron}, they fail to take these ideas to their logical conclusion. These models were not formulated in competitive-cooperative settings and did not create networks of independent agents. Instead, neurons are treated as performing Gibbs sampling of the posterior probability distribution to learn about stimulus features \cite{wang2015reinforcement}. Similarly, methods to formalize neurons as agents focused on the distribution of rewards via trusted third parties and auction theory to describe the multi-agent paradigm \cite{ohsawa2018neuron}.

Recently a neural model composed of multiple agents was proposed \cite{chalk2020training} to infer an objective function (inverse reinforcement learning) from network dynamics. Additionally, each neuron maximizes a reward function defined by the global network dynamics. The authors correctly identify the issue of instilling top-down solutions in order to model network dynamics. While quantifying and assessing the dynamics are important, the ideas presented in this paper focus on the behavior that arises from those dynamics. The primary differentiator is that, in this paper, neurons maximize their own self-interest, which is not a direct function of the state of all neurons in the network (see equation 5 in \cite{chalk2020training}). This formulation is grounded in local rewards derived from computational-biological principles.

Of particular interest to this paper is the work focused on unifying the fields of machine learning and neuroscience. Specifically topics like spiking neural networks \cite{diehl2015unsupervised}, biologically plausible backpropagation \cite{lillicrap2016random, baldi2018recirc}, and bio-inspired architectures \cite{ott2019learning, bengio2015towards}. These solutions offer algorithmic theories grounded in biologically accurate implementations. Further, there have been numerous calls to integrate neuroscience into machine learning \cite{hassabis2017neuroscience, marblestone2016toward, ott2019questions}.  Demand for this integration suggests that machine learning alone cannot offer a complete solution, and neuro-scientific principles may remedy some of deep learning's pitfalls \cite{marcus2018deep, su2019one}. 

\section{Neuron as an Agent}
\label{neurons}
\lettrine[lines=3]{\initfamily\textcolor{darkgreen}{N}}{eurons} in cortex are individual cells defined by a semi-permeable lipid bilayer. This membrane encapsulates the cell and isolates its contents from other neurons and the extracellular material. Neurons make connections to each other via synaptic junctions, where chemical or electrical information can be exchanged.

These individual actors compete with their neighbors for resources, inhibition, and activity expression. For neurons to remain healthy, they must receive adequate resources from the extracellular solution as well as homeostatic cells (i.e., glial cells). As in nature, resources may often be scarce, requiring trade-offs in their allocation. Therefore, neurons compete with each other to acquire these resources in order to survive. When activated, inhibitory neurons cause a decrease in the activity of their postsynaptic contacts. As a result, when excitatory neurons become active, they can trigger inhibitory interneurons, thereby reducing the activity of neighboring neurons. Indirect inhibition can also occur locally when one neuron takes necessary ions from the extracellular solution. Whether direct or indirect, inhibition leads to competition in expression, where one neuron becoming active results in its neighbor remaining silent. 

While competing with each other, neurons must simultaneously cooperate in order to code meaningful representations for the organism overall. To this end, we propose modeling neurons as reinforcement learning agents. Each neuron can be described by a policy, $\pi_{\theta_i}(a|o_t)$. Where the policy defines a distribution over actions, $a$, conditioned on an observation, $o_t$, of the environment at the current point in time. 

The observation, $o_t$, is received by the neuron via its incoming dendritic connections from other neurons. As no single neuron can observe the entire environment, the observation provides incomplete information. Any given neuron only has information about what its local neighbors have communicated to it through its synapses. Neurons signal each other by firing an action potential - spike - or remaining silent - no spike. This produces a binary action space, $a \in \{0, 1\}$. 

\begin{figure}
\centering 
\includegraphics[width=\textwidth]{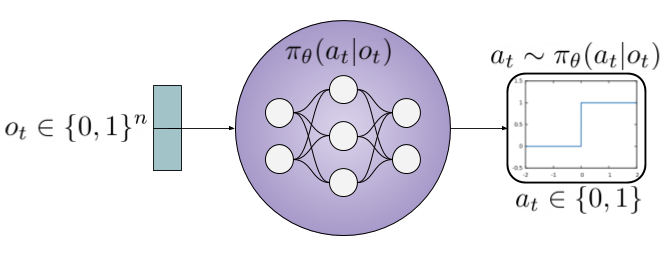}
\caption{\footnotesize Each neuron is independently described by its own policy, $\pi_\theta$. In this paper neuron's policies are parameterized by deep neural networks. The input to the neuron - in all non-input layers - is a binary vector. The output is a binary action, fire or remain silent.}
\label{fig:neuron}
\end{figure}

Policies may be described by a variety of approaches across the reinforcement learning literature. In this paper, we leverage stochastic policies trained by the policy gradient method. Stochastic policies sample an action according to the distribution over actions given by $\pi_\theta(a|s)$. The randomness induced by stochastic policies makes an appropriate connection to observations from biology.\footnote{Pyramidal neurons that fire an action potential, only release neurotransmitters into the synaptic cleft 20-50\% of the time}

It is essential to note the specific mathematical form of this policy is not important. The novelty of this approach comes via formulating cortical computation as a competitive-cooperative multi-agent reinforcement learning paradigm. This approach is in stark contrast to previous approaches across both computational neuroscience and machine learning. These fields handle reinforcement learning at the model level, where theories like temporal difference learning and reward prediction errors treat the network as an abstract entity. As a result, tasks are completed without regard for the network's constituents – neurons. 

\section{Multi-Agent Networks}
\begin{figure}
\centerline{\includegraphics[width=\linewidth]{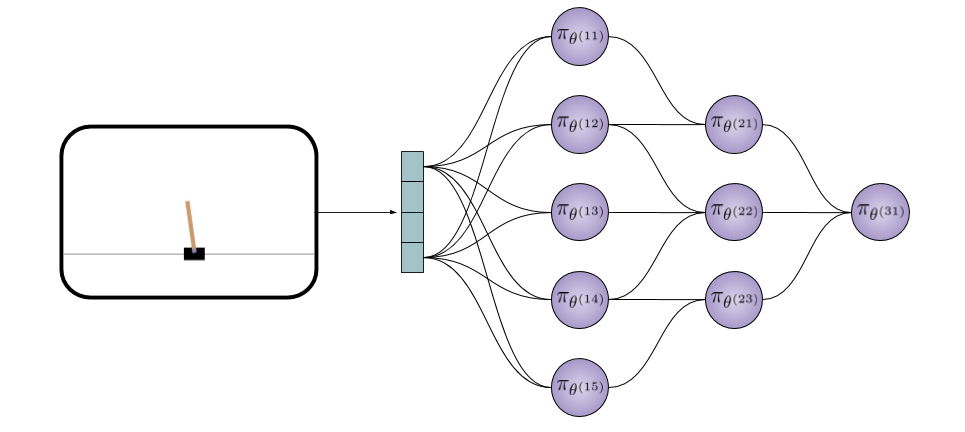}}
\caption{\footnotesize Network composed of independent neural agents. Each neuron maintains its own parameters describing its policy for taking actions. The first layer of the network receives input from the tasks and activity is passed to downstream connections. }
\label{fig:network}
\end{figure}

\lettrine[lines=3]{\initfamily\textcolor{darkgreen}{I}}{n order to} create networks of neurons, ideas presented in Section \ref{neurons} are replicated across a population. The population is wired together via directed connections to create a network. This construction yields a multi-agent reinforcement learning (MARL) paradigm. MARL defines a setting with multiple independent agents taking actions with incomplete information of the others. This can be framed as neurons in the population composing the environment of other neurons. Figure \ref{fig:network} depicts the multi-agent network. Each neuron maintains its own parameters describing its policy for taking actions. The first layer of the network receives input from the task environment, and activity is propagated downstream via directed connections. Observing the activity from the last neuron in the network provides a signal by which to take behavior-level actions in the task environment. 

\subsection{Learning in Multi-Agent Networks}
When learning is formulated in reinforcement or deep learning settings, it is in the context of minimizing cost functions. The minimization of a given function is correlated with performance on a behavior-level task like object detection. This becomes somewhat trivial, as gradient descent is explicitly designed to descend cost functions in parameter space. Iteratively moving to regions where those parameter settings yield lower values in the corresponding loss function. Unfortunately, there are severe issues with the approximations and abstractions these formulations assume \cite{ott2019questions, ott2020hayek}. 

Instead, the paradigm of multi-agent networks focuses attention on each individual neuron. Where every neuron learns via the policy gradient method, but the behavior of the network emerges out of these local interactions. This approach is in direct opposition to explicit minimizations of global cost functions via gradient descent. 

Each neuron is trained to maximize its own self-interest. In the following section, we describe the incentives provided to neurons that can be maximized. By relating the performance on high-level tasks to information locally available to neurons, they learn how to act so as to maximize rewards for the population. 

\section{Incentives for Multi-Agent Networks}
\label{incentives}

\lettrine[lines=3]{\initfamily\textcolor{darkgreen}{R}}{eward functions} are defined by the environment. They provide agents with a tangible signal in response to their actions. Agents can then modify their actions in order to achieve more rewards over time. In this section, we describe ways to incentivize networks of independent agents to act cooperatively. This is accomplished by a variety of reward schemes. Rather than explicitly defining how neurons are to act, we introduce incentives that guide the underlying dynamics of multi-agent networks. 

\subsection{Task Rewards}
The common approach defined by reinforcement learning is to maximize the expected reward over time \cite{sutton2018reinforcement}. This is accomplished by directly maximizing the task rewards given by the environment. In order to provide neurons with information about the high-level tasks, we distribute the reward information globally to all neurons in the population. For example, when organisms consume food, the population is rewarded for this successful endeavor. Similarly, neuronal populations, reward all constituents upon successful completion of a task.

The experiments presented in this paper leverage the OpenAI Gym for reinforcement learning tasks \cite{brockman2016openai}. In experiments run using the CartPole environment, a task reward is derived for balancing a pole vertically. At every time step, each neuron receives a reward corresponding to the networks high-level performance. However, if all agents simultaneously maximize only the task reward, there is no process to induce competition between neurons. In order to impose competition, in hopes of creating more sound dynamics, we introduce a variety of biologically motivated rewards described below.

\subsection{Biological Rewards}
Neurons may be rewarded based on a variety of events and interactions with other agents in their environment. Rewards proposed in this paper come from a variety of interactions. Neurons receive rewards via global task-based performance signals or local signals within neighborhoods of multi-agent networks. The rewards proposed in this subsection are derived from observations regarding the function of cortical circuits. 

\textbf{Activity:}
In order for neurons to communicate information to one another, they must become active. This formulation states the goal of neurons is to fire signals, thereby encoding information via their spikes. As a result, neurons that fire receive a positive reward, whereas neurons that do not receive a negative reward.

\textbf{Sparsity:} 
The amount of activity, or lack thereof, is an important aspect of cortical networks. Layers that are too sparse do not provide enough information and are highly susceptible to noise. If only a few neurons are active, the semantic meaning of these neurons can be changed easily by a few erroneously active neurons \cite{foldiak2003sparse}. However, if layers are not sparse enough, similar issues arise. A population with all neurons being active cannot code any meaningful information. Consequently, neurons optimize for a specific sparsity level. Within the population, neurons that become active are rewarded for meeting sparsity levels, and neurons that violate sparsity levels are penalized. 

\textbf{Prediction:}
In order to learn high-level stimuli, neurons must be temporally aware. This temporal awareness manifests itself by learning sequences. Instead of learning sequences at the network-level, we propose learning sequences within the network. In order to incentivize this, agents must identify which neurons will become active next. Consequently, neurons receive rewards when their activity predicts the activity of their postsynaptic contact. Conversely, neurons that erroneously predict inactive neurons receive penalties. 

\textbf{Activity Trace:}
The rewards described thus far lead to an obvious attractor in the network dynamics. The same group of neurons will always become active, thus achieving the correct sparsity level and correctly predicting the next active neuron downstream. In order to remedy this, we introduce an activity trace. This reward penalizes neurons that are always active or always silent. Biological neurons can only fire so often in a short period, as local resources and the health of the cell serve as activity constraints. Similarly, neurons that never fire are unlikely to be maintained and kept healthy. This reward enforces diversity in the coding of activity across the population.

Neurons must learn to balance these rewards, maximizing their own self-interest while making trade-offs for the good of the population. Neurons may want to become active at every time step based on the activity reward. However, this will lead to a violation of sparsity, activity trace, and a decrease in the task level reward. As a result, agents must learn to make trade-offs between these rewards. Ultimately, finding the optimal time to become active to acquire rewards while encoding useful information about the stimuli. With neurons, multi-agent networks, and reward schemes now introduced, we now describe settings for experiments conducted.

\section{Experiments}

\begin{table}
\centering
\begin{tabular}{l|ccc}
\toprule
{} & \multicolumn{3}{|c}{Number of Layers} \\
Reward &        1 &        2 &             5 \\
\midrule
All          &   1,480 &   \textbf{1,910} &   4,824 \\
Bio $\rightarrow$ All &   \textbf{1,275} &   3,718 &   \textbf{4,419} \\
Task         &   7,691 &  20,000 &  20,000 \\
\bottomrule
\end{tabular}
\caption{Average number of trials required to reach a one hundred step mean reward of 300 in CartPole. }
\label{tab:results}
\end{table}

The networks implemented in this preliminary study used the CartPole environment from the OpenAI gym \cite{openaigym}. Networks were trained with one, two, and five layers. The reward type for neurons consisted of only the high-level task reward, all rewards described in Section \ref{incentives}, or the biological rewards for 1,000 trials then all rewards. Each network was trained for a maximum of 20,000 episodes. Simulations were terminated once the network achieved an average reward of 300 or more, on the previous one hundred episodes. Each network configuration was repeated ten times to account for noise.

Table \ref{tab:results} displays the average number of episodes required to reach the stopping condition. As the number of layers increases, it is clear that the task reward alone becomes insufficient to train the network. These results indicate that the biological rewards serve an essential purpose. They are able to give the network better dynamics such that it can perform better on high-level tasks. As a result, training deeper networks are possible.

The current version of this paper presents very preliminary results on the CartPole environment. The demonstration serves to prove two points. First, the proposed multi-agent network is capable of learning a high-level task. Second, biologically motivated rewards aid in modulating dynamics for more extensive networks. 
Future versions of this paper, or follow up papers will explore deeper networks implemented on more reinforcement learning tasks. Future areas of exploration include different agent formulations (deep Q networks or actor-critic models) and different update schedules, like asynchronous methods. The code is publicly available at: \url{https://github.com/Multi-Agent-Networks/NaRLA}.

\section{Conclusion and Future Work}
\lettrine[lines=3]{\initfamily\textcolor{darkgreen}{M}}{any aspects} of neurological function have been modeled to date. These models often range in their attention to biological detail and application scalability. Ideas for computational models are often one-sided, focusing on the minutia or the very abstract (behavior-level loss functions). In this paper, we brought together these often disparate approaches. 

We introduced framing neurons as individual actors, each pursuing their own self-interest. We demonstrated a network composed of independent agents capable of cooperating in order to accomplish high-level tasks. Future work in this area should investigate bigger models, more complex tasks, and more local, neuron specific, rewards in order to provide network stability and smooth the learning process. 

The primary issue with reinforcement learning comes from its reliance on environment-specific reward functions. While this method has produced impressive results \cite{agostinelli2019solving, mnih2015human}, it relies on explicit reward functions defined external to the agent. This paradigm directly opposes biology, where the goodness of a stimulus is subjective to the organism and changes with experience \cite{ott2019questions}. Thus, task-level rewards are another imposed heuristic designed by researchers \cite{ott2020hayek}. Future work must investigate the production of rewards internal to the agent. By moving the locality of reward generation inside the agent, there is no longer a strict reliance for the agent to directly maximize the designated reward. Instead, it allows the agent to asses stimuli, assign a value to it, and take actions to achieve their intrinsic goals. 

\printbibliography
\end{document}